%% file: main.tex
\pdfoutput=1
\documentclass[runningheads]{llncs}

\usepackage{eccv}

\usepackage{eccvabbrv}

\usepackage{graphicx}
\usepackage{subcaption}       % subfigures (used by ECCV example as well)
\usepackage{booktabs}         % professional tables
\usepackage{amsmath}          % math environments
\usepackage{amssymb}          % for \checkmark
\usepackage{multirow}         % multi-row cells in tables
\usepackage{microtype}        % microtypography
\DeclareRobustCommand{\vec}[1]{\mathbf{#1}}

\usepackage[accsupp]{axessibility}

\usepackage{hyperref}

\usepackage{orcidlink}

\usepackage{pifont}
\usepackage{eurosym}

\usepackage{tikz}
\usetikzlibrary{positioning, shapes.geometric, arrows.meta, fit, backgrounds, calc}

\usepackage{tcolorbox}
\usepackage{listings}
\lstdefinelanguage{json}{
  basicstyle=\tiny\ttfamily,
  showstringspaces=false,
  breaklines=true,
  numbers=none,
  columns=fullflexible,
  morestring=[b]",
  stringstyle=\color{teal!70!black},
  keywordstyle=\color{blue!70!black},
  commentstyle=\color{gray},
  literate=
    {ä}{{\"a}}1
    {ö}{{\"o}}1
    {ü}{{\"u}}1
    {Ä}{{\"A}}1
    {Ö}{{\"O}}1
    {Ü}{{\"U}}1
    {ß}{{\ss}}1
    {é}{{\'e}}1
}
\usepackage{acronym}
\newacro{anls}[ANLS]{Average Normalized Levenshtein Similarity}
\newacro{cer}[CER]{Character Error Rate}
\newacro{vlm}[VLM]{Vision-Language Model}
\newacro{ocr}[OCR]{Optical Character Recognition}
\newacro{kie}[KIE]{Key Information Extraction}
\newacro{llm}[LLM]{Large Language Model}

\begin{document}
% %
\title{Lot Machine}
\subtitle{Multimodal Lot Extraction from Auction Catalogs}
\author{Mathias Zinnen\inst{1}\orcidID{0000-0003-4366-5216} \and
Alisha Mund\inst{1}\orcidID{0009-0000-0083-6412} \and
Sabine Lang\inst{2}\orcidID{0000-0003-2543-0085} \and
Lukas Hüttner\inst{1}\orcidID{0009-0001-0231-7517} \and
Thomas Gorges\inst{1}\orcidID{0009-0007-0573-0992} \and
Vincent Christlein\inst{1}\orcidID{0000-0003-0455-3799}
} % add potential contributors here (document analysis group and maybe gabi?)
\authorrunning{M. Zinnen et al.}
\institute{Pattern Recognition Lab, FAU Erlangen, Germany\and
Department of Digital Humanities and Social Studies, FAU Erlangen, Germany
}
\maketitle           

\begin{abstract}
For provenance research and art market studies, auction catalogs are an essential resource to trace specific objects over time and space.
While historical auction catalogs follow established domain conventions, their internal formatting remains highly variable, 
and their large-scale analysis is currently restricted by the lack of machine-readable representations of the auction lots.
We propose a pipeline to automatically extract structured lot-level metadata from German Sales, a large database of historical auction and sales catalogs from the 19th and 20th centuries.
Using a manually annotated test set of representative catalog pages,
we evaluate \acp{vlm} under varying prompt strategies and constrained decoding frameworks. 
To reflect the practical constraints faced by cultural heritage institutions, including budget, compute resources, and data privacy requirements, 
we benchmark the methods across different deployment modes ranging from commercial providers to locally hosted, quantized models.
We find that commercial endpoints establish the performance ceiling, while institutional gateways offer a viable, privacy-preserving alternative. 
Local deployments remain feasible, but strictly require enforcing the output structure during generation to guarantee a valid JSON format.
While varying degrees of human-in-the-loop correction are still necessary, this work demonstrates that a \ac{vlm}-based pipeline can successfully unlock historical auction catalogs for large-scale automated analysis.

\end{abstract}

\section{Introduction}
% Provenance Research / German Sales Context
% The aim of provenance research is to reconstruct the ownership history of objects over time and space. 
% It also studies how ownership changed, for example, through sale, inheritance, theft, or seizure. 
The aim of provenance research is to reconstruct an object's ownership history over time and space by examining changes in ownership, for example, through sale, inheritance, theft, or seizure. 
Documenting the transfer of ownership of an object allows researchers to establish its legal ownership and helps to determine its authenticity and authorship~\cite{yeide2001aam}. 
In Germany, provenance research encompasses different areas, including objects looted by the National Socialists, expropriations in the German Democratic Republic or Soviet occupation zone, and cultural property from colonial contexts. 
To reconstruct ownership changes, provenance researchers study different sources, such as the objects themselves, literature, archival sources, and online databases~\cite{baresel2019provenance}. 

An important online database for provenance research is German Sales.
Currently, German Sales contains over 15,500 digitized auction and sales catalogs, together with descriptive bibliographic metadata~\cite{germansales}. The catalogs mainly originate from German-speaking countries and were predominantly published between 1901 and 1945.
These catalogs not only contain information about objects offered during an auction but sometimes also hold information about prices, sellers, and buyers -- in the form of handwritten annotations. 
% Problem Description

% Currently, these catalogs are accessible via a text-based search, made possible through generic Optical Character Recognition (OCR). 
Currently, these catalogs are described by bibliographic metadata, findable through a persistent identifier and accessible via text-based search enabled by generic \ac{ocr}.
While full-text search provides a baseline for discovery, it is insufficient for automated, large-scale analysis.
The lack of standardized, structured records to represent the auction lots prevents researchers from systematically tracking spatial and temporal market trends, artist popularity, or conducting linguistic analyses of the object descriptions.
Extracting structured lot-level metadata would address these limitations while also enabling the German Sales catalogs to be integrated into Linked Open Data networks such as Wikidata and connected with other provenance resources, including online museum collections and the Lost Art Database~\cite{sattler2002projekt}.
%Extracting structured lot-level metadata would overcome these limitations.
%Furthermore, it would enable the integration of the catalogs into Linked Open Data networks such as Wikidata and allow for connecting the German Sales catalogs to other provenance databases such as online museum collections or the Lost Art Database~\cite{sattler2002projekt} to facilitate cross-database provenance tracing. 
\begin{figure*}[t]
    \centering
    % \fbox{
        \input{figs/overview3modes.tex}
        % \input{figs/overview.tex}
    % }
    \caption{Overview of the proposed Key Information Extraction pipeline.}
    \label{fig:pipeline_overview}
\end{figure*}
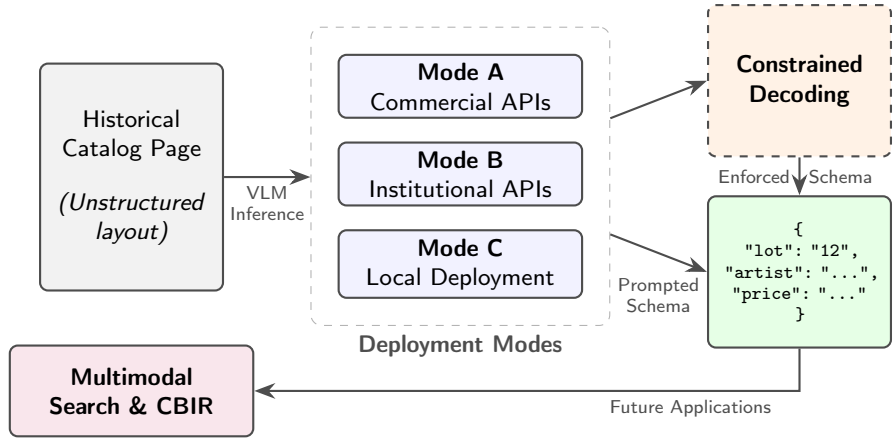

% Related Work
%% Introduce and concisely explain KIE, introduce benchmark FUNSD
The automated transition from unstructured images to structured metadata is referred to as \ac{kie}. 
With growing importance of the task, standardized benchmarks such as FUNSD (Form Understanding in Noisy Scanned Documents)~\cite{jaume2019funsd} 
enable benchmarking models with respect to text detection, layout analysis, and entity linking.
%% Traditional approaches: traditional OCR + layout parsing / rule-based assignment (explain and give examples)
Historically, \ac{kie} relied on decoupled, multi-stage pipelines, where the raw text is first extracted via an \ac{ocr} engine, 
and then grouped into logical key-value pairs via rule-based layout parsing methods such as hierarchical layout subdivision~\cite{nagy1992prototype} or clustering of page components~\cite{o1993document}.
While these approaches proved effective for highly standardized forms with a fixed layout, they fail on the German Sales catalogs due to their layout variability, published by multiple publishers over a period of over 100 years.
%% Hybrid: Traditional OCR + LLM-based text structuring 
To overcome the limitations of heuristic rules, recent hybrid approaches pipe the raw output of traditional \ac{ocr} engines into \acp{llm} to perform zero-shot text structuring and semantic parsing~\cite{stewart2025retrieving}. 
While this two-stage methodology successfully extracts structured data from historical texts, it introduces unnecessary complexity. 
It requires maintaining separate text detection, recognition, and language models, and frequently suffers from error propagation where the \ac{llm} loses visual layout context.
Newer approaches like olmOCR~\cite{poznanski2025olmocr} and DocVLM~\cite{nacson2025docvlm} address this by directly injecting \ac{ocr} output into \acp{vlm} but have not yet been applied to historical data.
%% Our approach: Direct structure generation
In this work, we bypass decoupled pipelines in favor of direct structure generation, minimizing system complexity by requiring only a single model to be deployed. 
Multimodal architectures such as LayoutLMv3~\cite{huang2022layoutlmv3} and, more recently, end-to-end \acp{vlm} like Qwen-VL~\cite{bai2023qwen}, Gemma~\cite{team2024gemma}, 
or InternVL~\cite{wang2025internvl3} can autoregressively generate structured output directly from the pixel input.
Current end-to-end \acp{vlm} generally employ varying architectural strategies to process this visual input. 
Models like LLaVa~\cite{liu2023visual} or PaliGemma~\cite{steiner2024paligemma} rely on dense architectures that pair highly compact vision encoders with massive language decoders, forcing the model to resolve spatial layouts through text-based contextual reasoning. 
Conversely, InternVL~\cite{wang2025internvl3} and variants~\cite{luo2025mono,gao2024mini} rely on larger vision backbones, which enable precise visual feature extraction before the text decoder is engaged.
Furthermore, to mitigate the computational burden of these scaled architectures, many state-of-the-art \acp{vlm} now integrate Mixture of Experts (MoE) routing. 
MoE models like Mixtral~\cite{jiang2024mixtral} or, more recently, MoE-LLava~\cite{lin2026moe} activate only a small, highly efficient subset of parameters during inference, which allows higher overall parameter counts while maintaining efficiency and accuracy.

% However, integrating their generative outputs into databases requires strict adherence to fixed domain schemas. 
To prevent structural hallucinations, recent frameworks employ Logit-level constrained decoding~\cite{willard2023efficient}, which uses a Finite State Machine to deterministically guide the \ac{vlm} into generating perfectly formatted JSON.

However, transitioning from visual layouts to structured schemas is not merely a technical challenge, but a semantic one.
Historical auction catalogs frequently feature variable formatting, implicit cross-references, and subjective descriptors (e.g., blending stylistic epochs like "Barock" with physical descriptions). 
Consequently, evaluating extraction fidelity requires not only robust architectures but also a critical examination of the metrics used to score them against inherently ambiguous humanities data.

% Proposed Solution / Contributions
While these technological advances, alongside the legacy of specialized transcription tools like Transkribus~\cite{kahle2017transkribus} or eScriptorium~\cite{kiessling2019escriptorium}, offer powerful theoretical solutions,
cultural heritage institutions face significant barriers to their adoption.
The deployment of \acp{vlm} involves a complex trade-off between extraction accuracy, hardware availability, technical expertise, budgetary limits, and data privacy requirements.
Relying on commercial APIs poses data sovereignty risks for unpublished archival material and creates dependencies on the goodwill of commercial institutions, their pricing policies, and terms and conditions.
Hosting open-weight models, on the other hand, demands substantial computational infrastructure and technical expertise.

To bridge the gap between theoretical capabilities and practical application, we designed and implemented an end-to-end \ac{vlm} extraction pipeline.
Using this implementation and a newly introduced, manually annotated benchmark of historical auction lots, 
we present a comprehensive empirical evaluation of \ac{vlm} deployment strategies for historical \ac{kie}. 
We systematically evaluate models across the entire deployment spectrum (summarized in \cref{fig:pipeline_overview}), 
ranging from commercial cloud APIs (Mode A) to publicly funded institutional gateways (Mode B) and quantized, locally hosted edge models (Mode C). 
% By standardizing the output generation through Logit-level constrained decoding, we isolate and measure the true visual-semantic capabilities of each approach.
Beyond infrastructural benchmarks, we conduct a field-level analysis across different catalog types to critically evaluate how standard structural metrics (\acs{anls}*, a structure-aware similarity score for hierarchical dictionaries) capture the nuances of subjective historical text compared to semantic overlap metrics (\textsc{rouge}-1).
From these empirical trade-offs, we derive actionable deployment guidelines, providing a pragmatic framework for archival institutions to select the optimal digitization pipeline given their specific real-world constraints.
The complete pipeline implementation, including benchmark data, deployment scripts and prompt templates, is publicly available\footnote{\url{https://github.com/mathiaszinnen/auction-lot-extraction}}.

\section{Methods \& Materials}
\subsection{The German Sales Dataset and Target Schema}
\label{sec:dataset}

\newcommand{\panelwidth}{0.190\textwidth}
\newcommand{\pageheight}{3.7cm}
\begin{figure*}[htbp]
    \centering
    \begin{subfigure}[b]{\panelwidth}
        \centering
        \includegraphics[width=\linewidth,height=\pageheight]{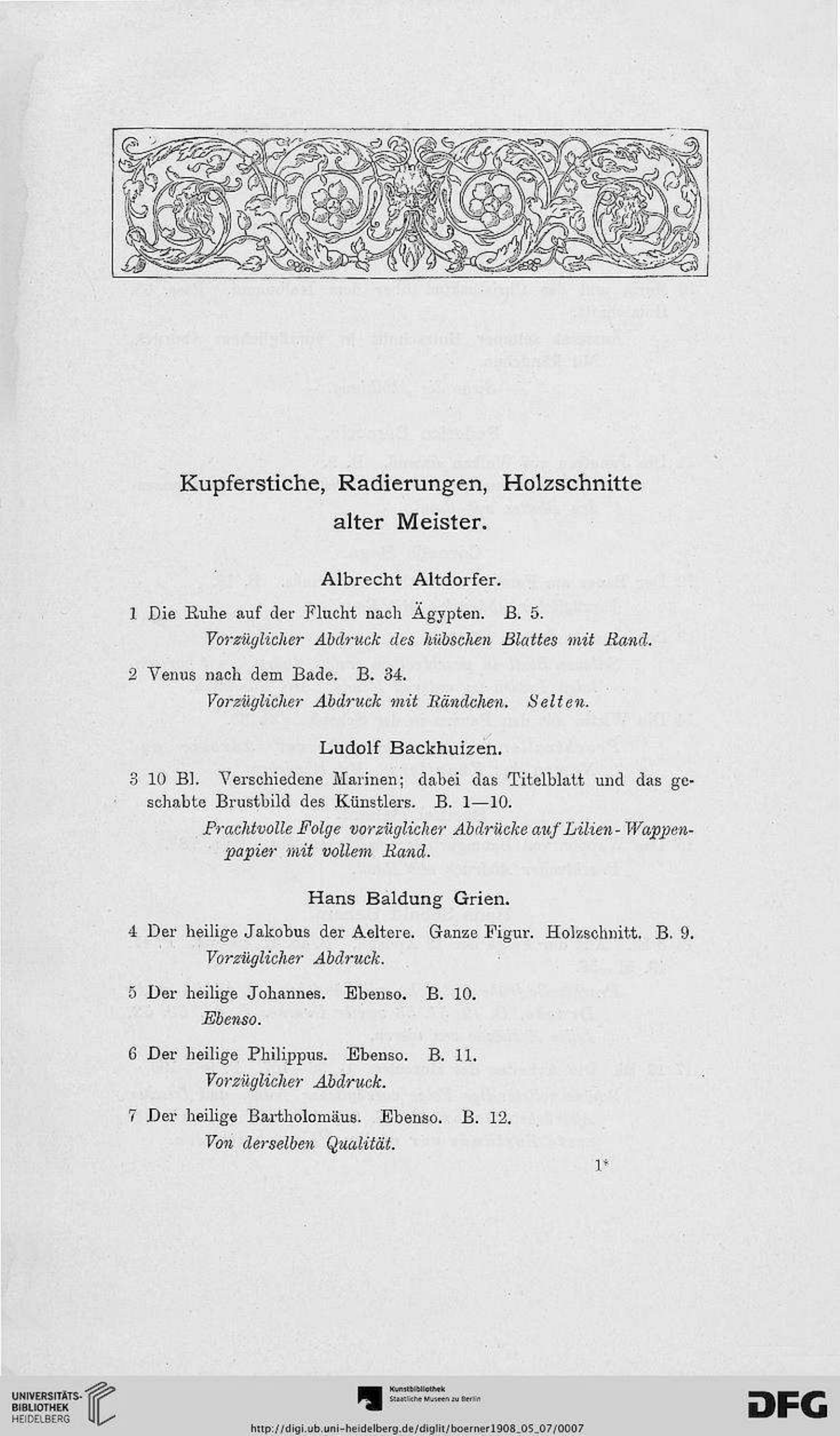}
        \subcaption{1908 (art)}
        \label{fig:cat1908}
    \end{subfigure}\hfill%
    % Panel B: 1909 Nöhring
    \begin{subfigure}[b]{\panelwidth}
        \centering
        \includegraphics[width=\linewidth,height=\pageheight]{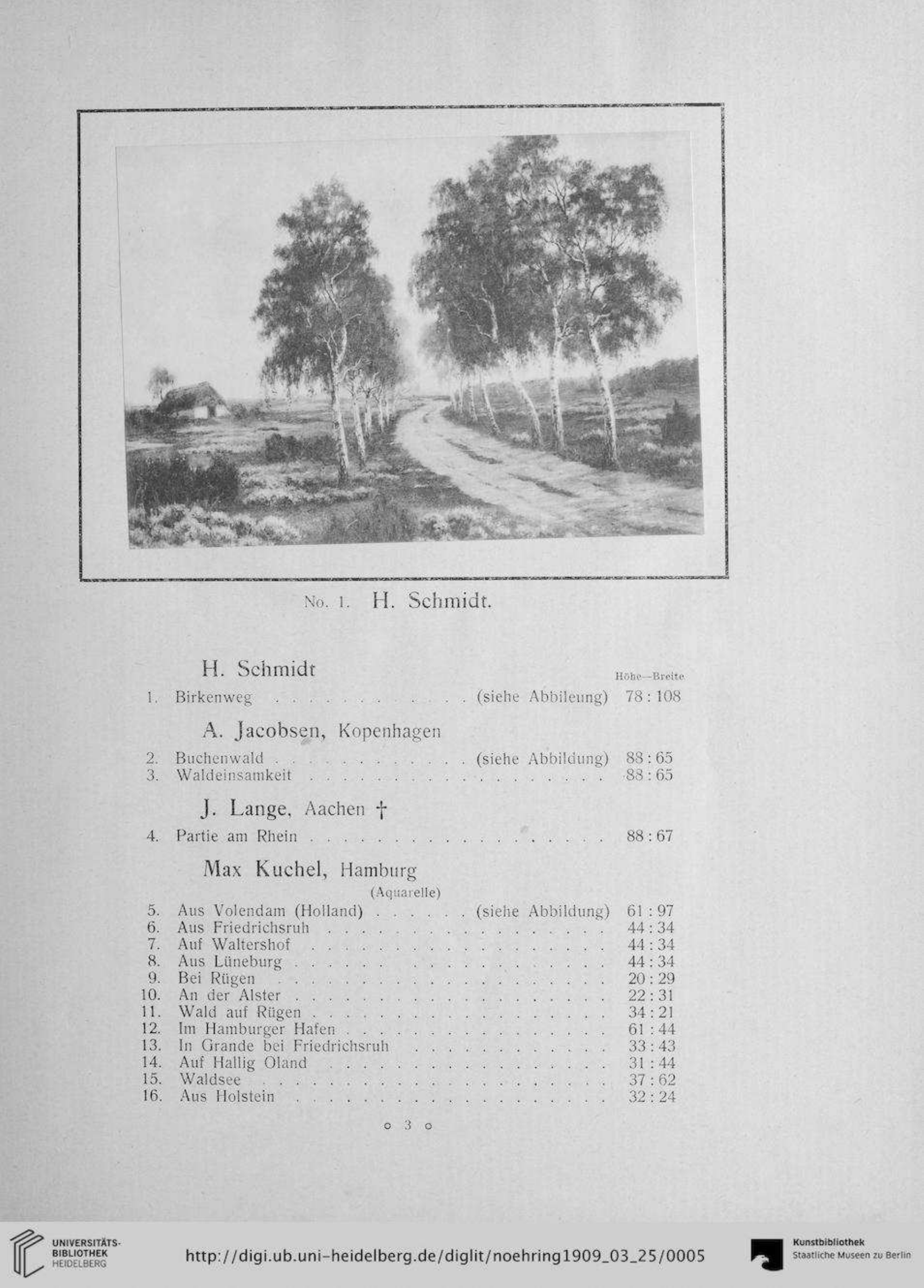}
        \subcaption{1909 (art)}
        \label{fig:cat1909}
    \end{subfigure}\hfill%
    % Panel C: 1931 Meilinger
    \begin{subfigure}[b]{\panelwidth}
        \centering
        \includegraphics[width=\linewidth,height=\pageheight]{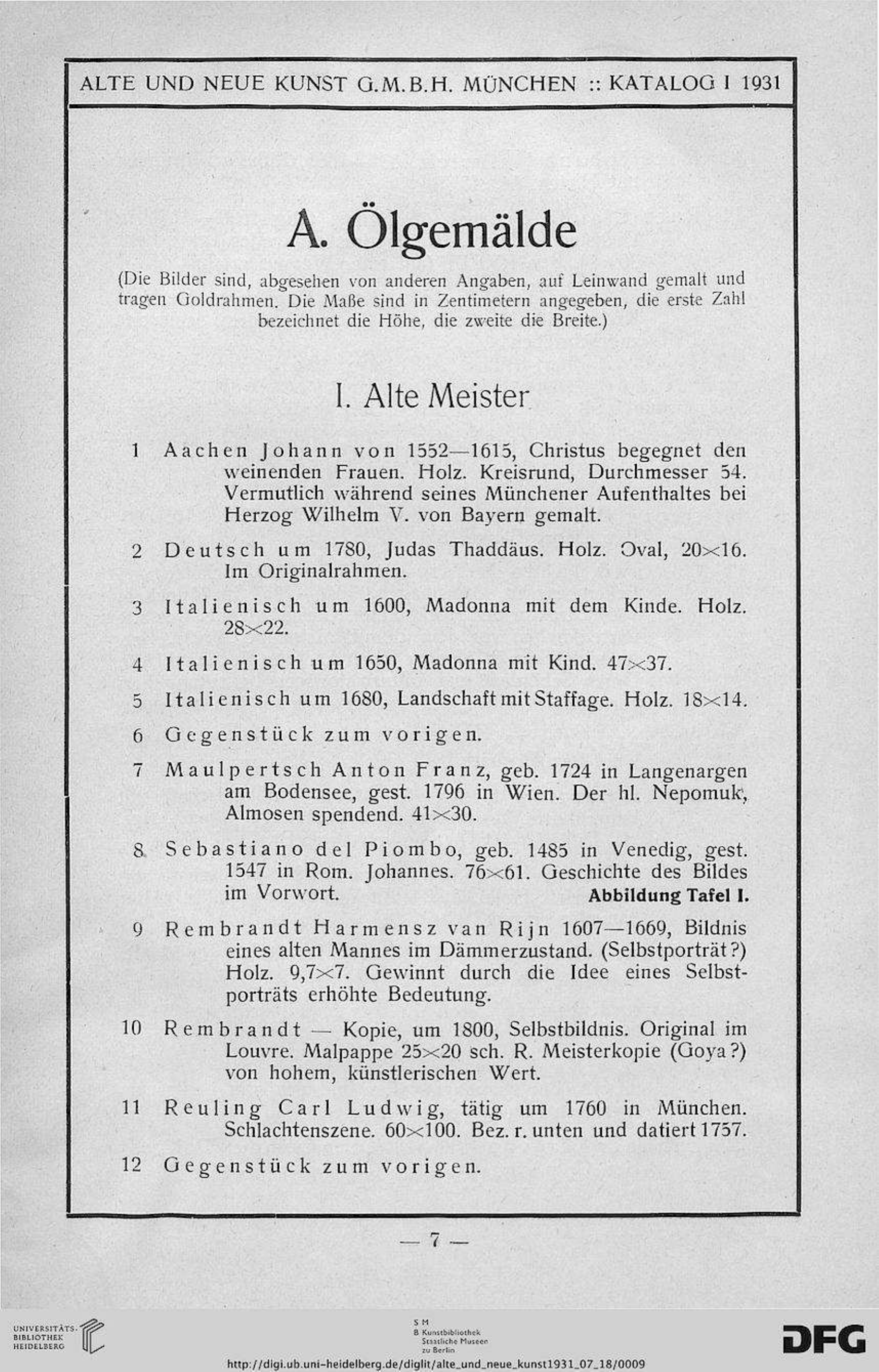}
        \subcaption{1931 (art)}
        \label{fig:cat1931}
    \end{subfigure}\hfill%
    % Panel D: 1932 Nagel
    \begin{subfigure}[b]{\panelwidth}
        \centering
        \includegraphics[width=\linewidth,height=\pageheight]{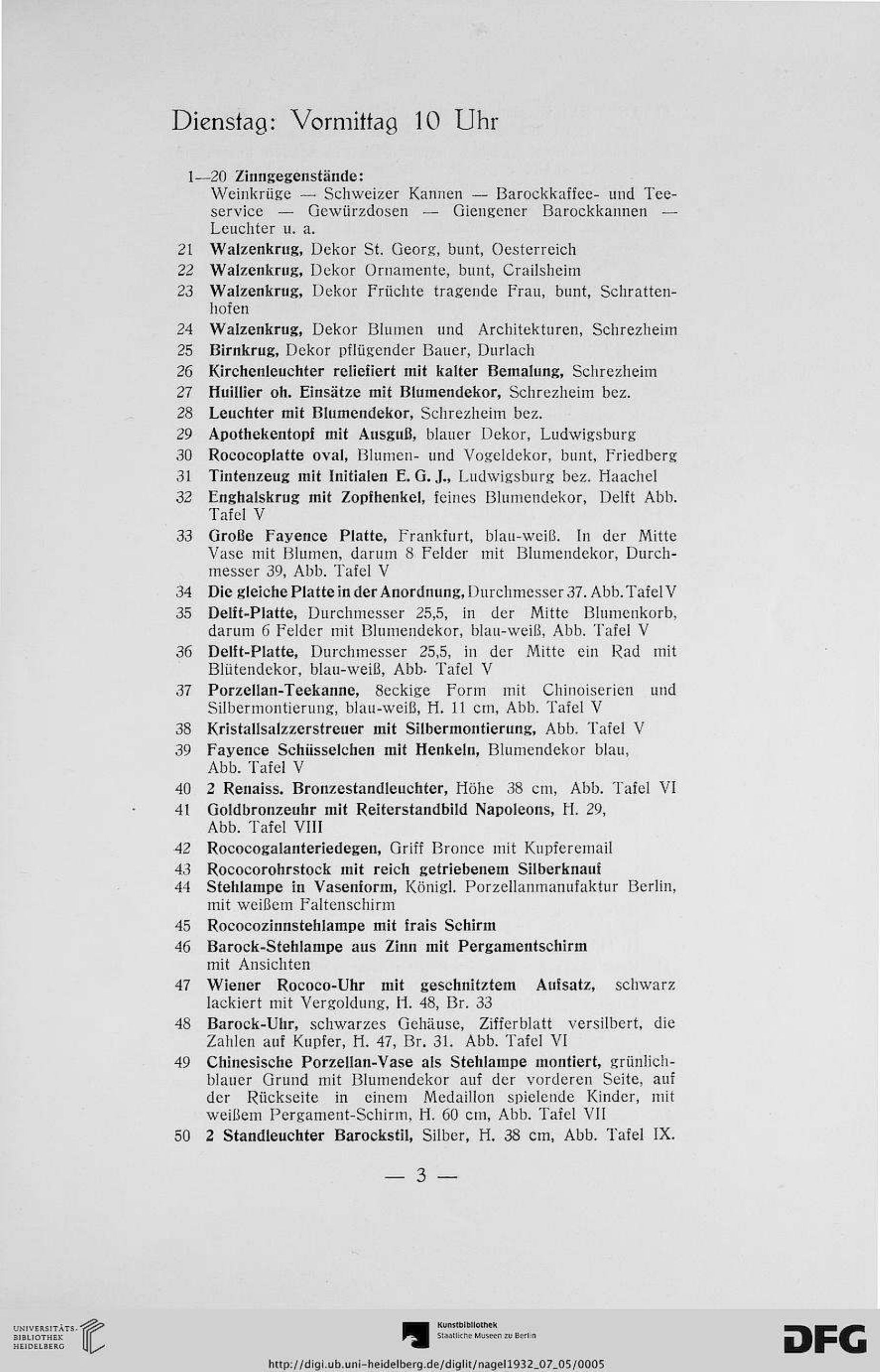}
        \subcaption{1932 (mixed)}
        \label{fig:cat1932}
    \end{subfigure}\hfill%
    % Panel E: 1935 AFAG
    \begin{subfigure}[b]{\panelwidth}
        \centering
        \includegraphics[width=\linewidth,height=\pageheight]{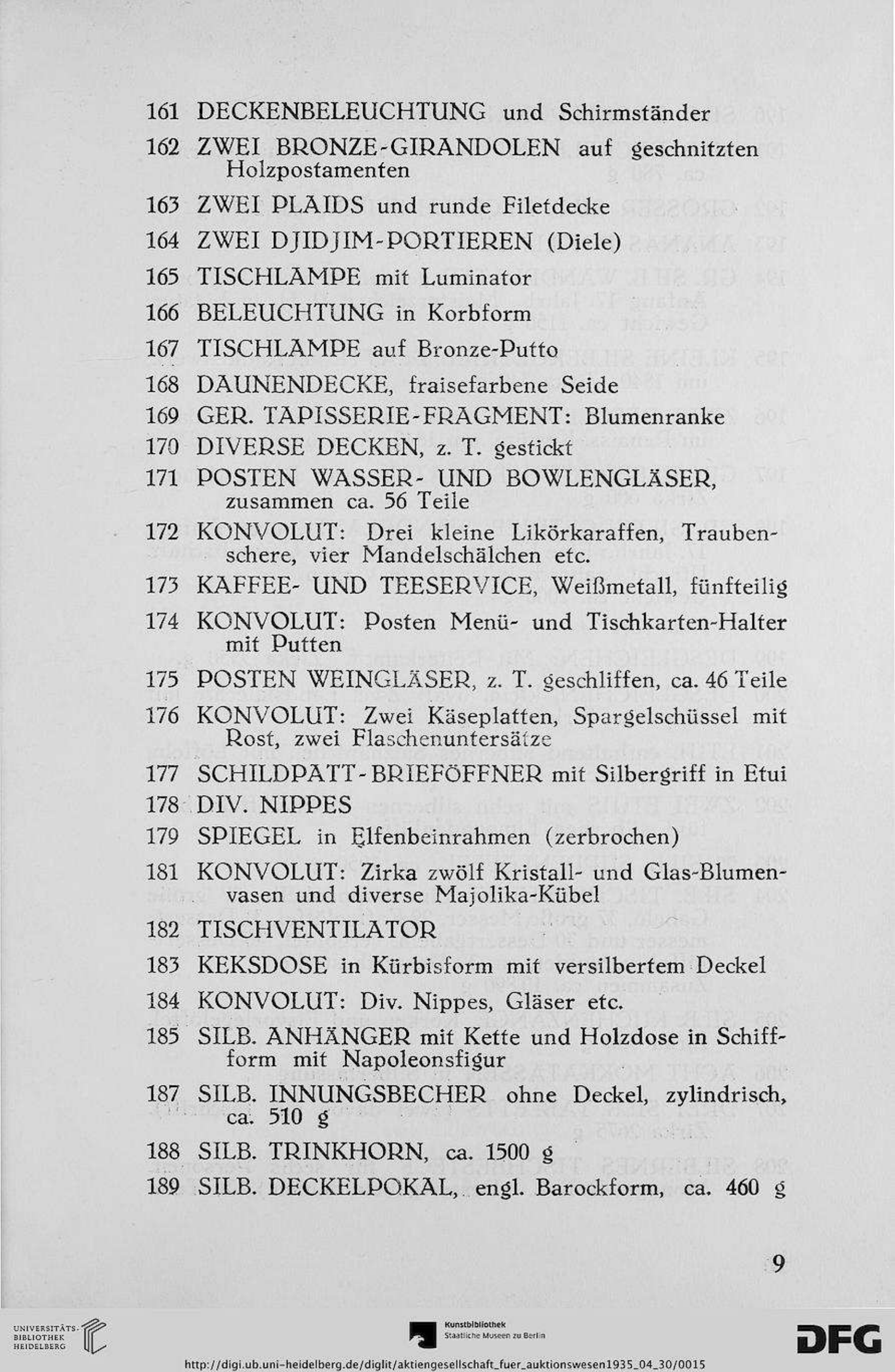}
        \subcaption{1935 (mixed)}
        \label{fig:cat1935}
    \end{subfigure}
    % \caption{Overview of historical auction catalog layout diversity across the test corpus. 
    % The figure displays representative pages from the five evaluated catalogs, illustrating the visual variations in object types, layout type and formatting.}
    \caption{Overview of the diversity in our test corpus. The figure displays representative pages from the five evaluated auction catalogs and illustrates visual variations regarding format and object type. Pages marked with "art" contain lots that would traditionally be considered "artworks", such as paintings, drawings, prints, and sculptures. The category "mixed" also includes objects such as furniture or vases. Catalog samples from~\cite{germansales}. }
    \label{fig:catalog_overview}
\end{figure*}

% For our experiments, we selected a representative subset of five catalogs from the German Sales database~\cite{cat1908,cat1909,cat1931,cat1932,cat1935}. 
% They span a range from 1908 to 1935 and present a large variety of types of objects offered through the lots. 
% To maintain readability, we refer to these catalogs by their publication years. 
% Two catalogs (1909, 1931) contain paintings, one catalog (1908) has copper engravings and woodcuts, two (1932, 1935) offer mixed objects such as kitchenware or furniture.
% \Cref{fig:catalog_overview} displays representative pages from the five evaluated catalogs, illustrating the visual variations in publication context, object type, and formatting.
% While structured lots related to titled artworks such as paintings, kupferstiche and holzschnitte are arguably most relevant for provenance research, 
% we still decided to also cover the more difficult and nonstandard objects found in the mixed object catalogs because this allows for analyses of market trends and further analyses beyond canonic art historic objects.
% \Cref{fig:dataset_examples} shows two exemplary lots next to their manually annotated target schema. 

\begin{figure}[t]
    \centering
    % kunst
    \includegraphics[width=0.9\linewidth]{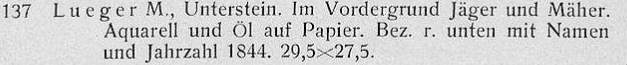} 
    
    \vspace{0.2cm}

    % stuhl
    \includegraphics[width=0.9\linewidth]{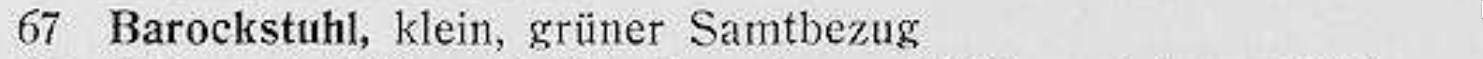}

    \begin{minipage}[t]{0.48\textwidth}
\begin{lstlisting}[language=json]
    {
        "lot_number": "137",
        "raw_text": "Lueger M., Unterstein. [...] 29,5x27,5.",
        "creator": "M. Lueger",
        "object_type": "Aquarell und Ölgemälde",
        "object_title": "Unterstein",
        "place_of_creation": "",
        "creation_time": "1844",
        "dimensions": "29,5x27,5",
        "height": "29,5",
        "width": "27,5",
        "depth": "",
        "weight": "",
        "description": "Im Vordergrund Jäger und Mäher, Bez. r. unten mit Namen und Jahrzahl"
    }
\end{lstlisting}
    \end{minipage}
    \hfill % Pushes the code blocks apart
    \begin{minipage}[t]{0.48\textwidth}
    \begin{lstlisting}[language=json]
    {
        "lot_number": "67",
        "raw_text": "Barockstuhl, klein, grüner Samtbezug",
        "creator": "",
        "object_type": "Stuhl",
        "object_title": "",
        "place_of_creation": "",
        "creation_time": "",
        "dimensions": "",
        "height": "",
        "width": "",
        "depth": "",
        "weight": "",
        "description": "Barock, klein, grüner Samtbezug"
    }
    \end{lstlisting}
    \end{minipage}

    \caption{Representative lots from the German Sales test set alongside their manually annotated target schemas. 
    \textbf{Top:} Crops of lots offering an artwork (top) and a chair (bottom). 
    \textbf{Bottom left:} The unambiguous extraction for the artwork. 
    \textbf{Bottom right:} The extraction for the furniture object.
    The assignment of subjective descriptors like "Barock" to \texttt{description} rather than \texttt{creation\_time} highlights the inherent annotation bias required for non-standard items. Lots cropped from \cite{cat1931} and \cite{cat1932}, respectively.}
    \label{fig:dataset_examples}
\end{figure}

To evaluate the models, we introduce a new benchmark dataset comprising 1,378 manually annotated lots across 152 pages from 5 representative catalogs. 
Examples are shown in \cref{fig:catalog_overview}. 
To accelerate the annotation process, we generated initial baseline predictions using Mistral-OCR, which demonstrated the highest qualitative accuracy in preliminary tests.
We then conducted a rigorous review of every lot and a systematic rule-based correction to mitigate evaluation bias.

This bootstrapping approach has a direct consequence for the interpretation of our results:
Mistral-OCR both served as a candidate generator for the ground truth and achieved the highest score in our evaluation (\cref{tab:macro_performance}).
Although every lot was reviewed and corrected, an annotation bias towards the model's implicit structuring conventions cannot be ruled out.
The \ac{anls}* we report for Mistral-OCR should therefore be interpreted as an optimistic upper bound for the model's performance, rather than a definitive measure of its extraction fidelity.
The scores of all other models are unaffected, since none of them contributed to the annotations.
A second constraint concerns the absence of an inter-annotator agreement study. 
The complete corpus was annotated in one pass by a single annotator, so we cannot quantify the degree of subjectivity in the schema assignment.
 
Accordingly, the inherent subjectivity of structuring historical data remains a critical factor in our evaluation. 
This is especially true for non-standard objects, such as the furniture example in \cref{fig:dataset_examples}. 
For instance, determining whether the term "Barock" (Baroque) should be categorized as a physical description, a date of origin, or as part of the object's title is often semantically ambiguous. 

The complete annotated benchmark is available via the projects' GitHub repository and is archived on Zenodo\footnote{\url{https://zenodo.org/records/21933459}}.

\subsection{Model Selection and Experimental Design}
\label{sec:selection}
The experimental space of Vision-Language Models spans multiple deployment modes, provider ecosystems, architectures, and parameter scales. 
To keep the experiments tractable, we select a subset of models that represent specific institutional constraints, hardware limits, and data sovereignty requirements.

To establish an upper bound for extraction fidelity using unconstrained compute resources, we evaluate commercial cloud APIs (Mode A). 
We select Google's Gemini family, specifically comparing the lightweight, cost-optimized Flash variant against the computationally heavy Pro model. 
This comparison allows us to empirically determine whether the financial premium of flagship models results in improved accuracy. 
Alongside Gemini, we include Mistral-OCR. 
As a European-developed model, Mistral explicitly addresses the political and infrastructural priorities of European cultural heritage institutions regarding regional data processing and privacy.

Publicly funded institutional gateways (Mode B) offer a compelling alternative for institutions that want managed API access to large \acp{vlm} without sacrificing data sovereignty.
While we expect accuracy to be slightly lower, these platforms ensure strict data privacy and protect users from volatile commercial pricing or changing terms of service. 
Beyond gateways provided by universities to their own members, public providers such as AcademicCloud~\cite{doosthosseini2026saia} and the Helmholtz Cloud~\cite{strube2024helmholtz} service a broad network of partnering institutions.
Among these, we select AcademicCloud due to its large number of available multimodal models and general accessibility.

% Here we operationalize the architectural paradigms introduced previously. %todo rephrase transition maybe
Within AcademicCloud, we evaluate two architectural paradigms:
First, to evaluate the highly efficient, vision-centric approach, we select a Mixture of Experts (MoE) model from the InternVL family (specifically a 30B total parameter with 3B active parameters per token). 
This model leverages its large vision backbone to extract dense historical typography while maintaining rapid inference. 
Second, to evaluate the contrasting dense, language-centric paradigm, we select the Gemma family to test the viability of resolving spatial layouts primarily through text-based reasoning.
Initially, we planned to evaluate Qwen3.6 and Qwen3.5 MoE variants available at AcademicCloud, but excluded these models from the final experiments due to persistent timeout errors.

Finally, to represent institutions operating under budget and hardware restrictions, we evaluate two locally hosted edge computing environments via llama.cpp~\cite{llamacpp} 
(Mode C): An 8-bit quantized 8 billion parameter InternVL3 model (InternVL-8B-UD-Q8\_K\_XL) and a mixture of experts variant of Qwen3.6-35B with 3 billion active parameters (Qwen3.6-35B-A3B-MXFP4\_MOE). 
Both are provided by Unsloth via HuggingFace.
While this deployment mode requires greater technical expertise to maintain, it provides absolute control over data flow, privacy, and local deployment logic, enabling the seamless integration of custom logit-level constrained decoding frameworks. 
To systematically evaluate hardware limitations, we conducted an initial analysis of baseline inference latency using a CPU-only environment, 
which yielded an estimated execution time of over 12h for the test set. 
Consequently, we transitioned the local evaluations to GPU-accelerated environments, where we use a consumer-level NVIDIA Quadro RTX6000 with 24GB graphics memory.

\subsection{Prompting and Schema Adherence}

To ensure the Vision-Language Models generate valid JSON outputs that strictly align with our target schema, we employ a dual strategy: instruction-based prompting and system-level constrained decoding.

\begin{figure}[t]
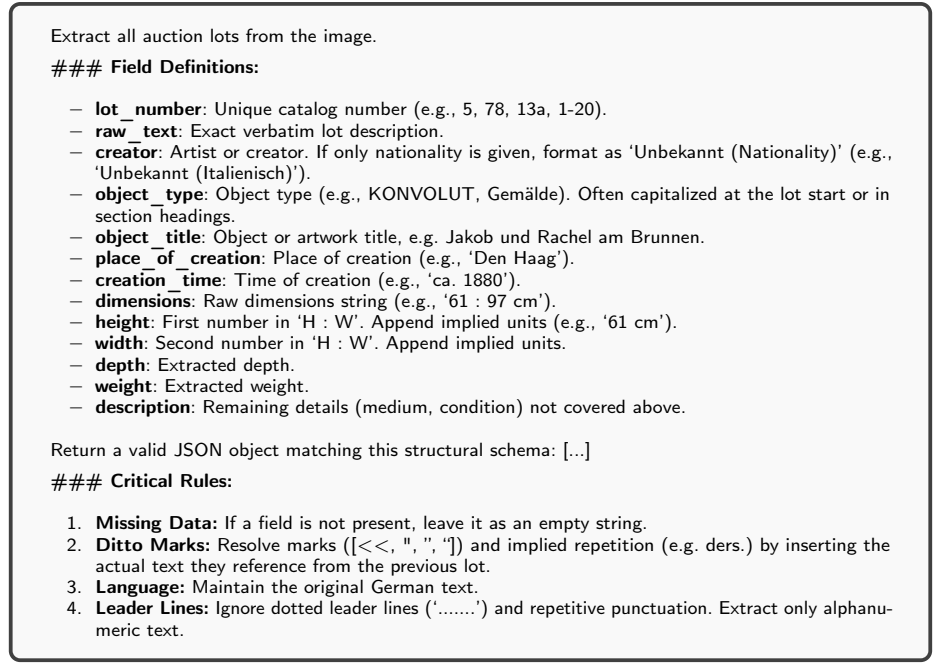

\begin{tcolorbox}[
    colback=gray!5!white, 
    colframe=gray!50!black, 
    fonttitle=\bfseries\sffamily, 
    fontupper=\scriptsize\sffamily,
]
Extract all auction lots from the image. 

\vspace{0.5em}
\textbf{\#\#\# Field Definitions:}
\begin{itemize}
    \setlength{\itemsep}{0pt}
    \setlength{\parskip}{0pt}
    \item \textbf{lot\_number}: Unique catalog number (e.g., 5, 78, 13a, 1-20).
    \item \textbf{raw\_text}: Exact verbatim lot description.
    \item \textbf{creator}: Artist or creator. If only nationality is given, format as `Unbekannt (Nationality)' (e.g., `Unbekannt (Italienisch)').
    \item \textbf{object\_type}: Object type (e.g., KONVOLUT, Gemälde). Often capitalized at the lot start or in section headings.
    \item \textbf{object\_title}: Object or artwork title, e.g. Jakob und Rachel am Brunnen.
    \item \textbf{place\_of\_creation}: Place of creation (e.g., `Den Haag').
    \item \textbf{creation\_time}: Time of creation (e.g., `ca. 1880').
    \item \textbf{dimensions}: Raw dimensions string (e.g., `61 : 97 cm').
    \item \textbf{height}: First number in `H : W'. Append implied units (e.g., `61 cm').
    \item \textbf{width}: Second number in `H : W'. Append implied units.
    \item \textbf{depth}: Extracted depth.
    \item \textbf{weight}: Extracted weight.
    \item \textbf{description}: Remaining details (medium, condition) not covered above.
\end{itemize}

Return a valid JSON object matching this structural schema:
[...]

\vspace{0.5em}
\textbf{\#\#\# Critical Rules:}
\begin{enumerate}
    \setlength{\itemsep}{0pt}
    \setlength{\parskip}{0pt}
    \item \textbf{Missing Data:} If a field is not present, leave it as an empty string.
    \item \textbf{Ditto Marks:} Resolve marks ([$<<$, ", '', ``]) and implied repetition (e.g. ders.) by inserting the actual text they reference from the previous lot.
    \item \textbf{Language:} Maintain the original German text.
    \item \textbf{Leader Lines:} Ignore dotted leader lines (`.......') and repetitive punctuation. Extract only alphanumeric text.
\end{enumerate}
\end{tcolorbox}
\caption{System prompt for instruction-based schema compliance.}
\label{fig:system_prompt}
\end{figure}

First, we guide the models' semantic focus using a standardized system prompt shown in \cref{fig:system_prompt}. 
% Note that the schema definition is omitted for brevity (marked by ``[...]'').
A critical component is the instruction to ignore typographical layout artifacts, specifically dotted lines (often used as visual leaders connecting text to prices or dimensions). During preliminary evaluations, we discovered that these repeating character sequences caused the Gemini models to get caught in infinite generative loops, leading to timeout errors and failed API calls. 
% To mitigate this, the core prompt provided to all models is structured as shown in \cref{fig:system_prompt}.

% These contain the data fields introduced in \cref{sec:dataset} along with a short description per field, including examples.

To enforce strict schema compliance on logit-level we additionally implement a constrained decoding approach, where we directly pass the target schema instead of including it in the prompt.
% Since the schema fields and descriptions are passed via the schema object, we remove the field definitions and schema definition in the prompt for this approach.

The technical implementation of constrained decoding varied across our deployment environments:
In Mistral, \ac{ocr} schema enforcement is handled natively through its built-in document understanding framework, where the target structure is defined and passed directly as Pydantic objects.
For Gemini, the AcademicCloud and locally hosted models, we enforce the structure using the standard \texttt{json\_schema} argument.
% The locally hosted models operate in principle like standard OpenAI API endpoints, but some adaptations were necessary:
% The schema definition had to be injected via the \texttt{extra\_body} field rather than the standard schema argument. 
% Furthermore, the schema dictionary itself required the explicit inclusion of the \texttt{"additionalProperties": false} flag to successfully lock the generation and prevent the gateway models from hallucinating keys outside of our target specification.

\begin{figure}[t]
    \centering
    \includegraphics[width=\textwidth]{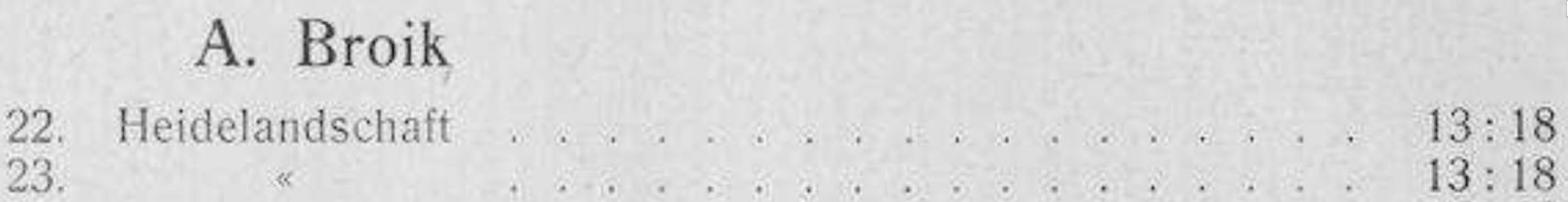}
    \caption{Example from the 1909 catalog~\cite{cat1909} illustrating the necessity of prompt rules to handle ditto marks and dotted leader lines.}
    \label{fig:dittodots}
\end{figure}

The example in \cref{fig:dittodots} illustrates why rules 2 and 4 are necessary: 
Resolving ditto marks is inherently complex and difficult to manage via rule-based post-processing, as these marks appear in variable formats that tend to be inconsistently transcribed by the models' \ac{ocr} recognition capabilities.
Explicitly addressing dotted leader lines was essential to stabilize the Gemini models, which otherwise became trapped in infinite decoding loops, endlessly generating repeating dots.
Finally, this example illustrates the challenge of forward references introduced by block headings, where an artist named at the top of a section implicitly applies to all subsequent lots.

% \subsection{Data Preprocessing and System Implementation}% can be skipped if space needed

% To prepare the raw archival PDFs for the models, we first programmatically skipped the library cover pages to isolate the actual catalogs. 
% We then used Ghostscript to convert all pages to single-channel grayscale and downsampled them to a uniform 150 DPI to reduce the PDF sizes.

% For the system architecture, we opted for standard real-time APIs over asynchronous Batch APIs to ensure the pipeline remains easy for other researchers to reproduce and run locally. 
% For Mistral-OCR, which operates on the level of complete documents, we used PyMuPDF to slice the optimized PDFs into two-page chunks directly in memory to prevent network time-outs.

\section{Results and Discussion}
In this section, we evaluate the proposed \ac{vlm}-based extraction pipeline, moving from a quantitative macro-analysis of computational and architectural performance to a qualitative examination of historical data complexities. 
We first establish the baseline infrastructural requirements, then compare state-of-the-art commercial APIs against privacy-preserving local deployments, and optimize the latter via hardware-aware ablation. Finally, we investigate how semantic ambiguities and cross-lot references in historical auction catalogs challenge current \ac{vlm} architectures, highlighting the limits of page-wise processing.
The code used for evaluating the methods is publicly available.\footnote{\url{https://github.com/mathiaszinnen/auction-lot-extraction/blob/main/evaluate.py}}

\subsection{Overall Comparison of Deployment Modes}

To evaluate the overall pipeline performance, we report three complementary metrics that capture structural accuracy, transcription fidelity, and computational feasibility for large-scale archival digitization. 

To quantify structural extraction fidelity, we employ \acs{anls}* \cite{peer2024anls}, a recent variant of the \acf{anls}. 
\ac{anls}* is specifically designed for \ac{kie} tasks and evaluates JSON outputs by penalizing both structural deviations (missing or hallucinated keys) and value-level inaccuracies.

To isolate the models' fundamental reading capabilities from their structural parsing, we evaluate raw transcription fidelity using the \ac{cer}.

For our quantitative evaluation, both \ac{anls}* and \ac{cer} are computed exclusively on lots matched between ground truth and detections via the lot number. 
Note that this excludes hallucinated or implicitly inferred lots from the primary penalty calculations. 
We decided on this approach because these structural over-generations can be filtered via deterministic post-processing (e.g., by filtering out lots without numbers assigned), penalizing them within the primary metrics would artificially deflate the models' actual transcription and schema-mapping capabilities. 
An evaluation including unmatched lot pairs penalizing hallucinated lot generation can be found in section S1 of the supplementary material.

Finally, we report the average inference time in seconds per page (sec/p) to contextualize the viability of deploying these models at an institutional scale.

% \subsection{Commercial APIs vs. Privacy-Preserving Local Deployment}
\begin{table}[t]
\centering
\caption{Macro-level comparison of the highest-performing configurations across deployment paradigms. Modes represent commercial cloud APIs (A), institutional gateways (B), and local edge deployments (C). Metrics include structural accuracy (\ac{anls}*), transcription fidelity (\ac{cer}), 
%financial cost per page (cost/p), 
and latency in seconds per page (sec/p).}
\label{tab:macro_performance}
\small
\begin{tabular}{clccc}
\toprule
Mode & Model &   \ac{anls}* ($\uparrow$) & \ac{cer} ($\downarrow$) & sec/p ($\downarrow$) \\
\midrule
\multirow{2}{*}{A} & Gemini-Flash  & 0.75 & 0.10 & 19.81 \\
& Mistral-OCR  & \textbf{0.87}  & \textbf{0.03} & 30.40 \\
\midrule
\multirow{2}{*}{B} & Gemma4-31B & 0.77 & 0.13 & 159.34 \\
& InternVL3.5-30B-A3B & 0.71 & 0.21 & \textbf{17.30} \\
\midrule
% C & InternVL-Q8 & 0.61 & 0.24 & 68.3 \\
\multirow{2}{*}{C} & InternVL3-Q8 & 0.61 & 0.24 & 68.30 \\
% & Gemma4-31B-Q4 &  \\
& Qwen3.6-35B-A3B & 0.72 & 0.16 & 78.96 \\ 
\bottomrule
\end{tabular}
\end{table}

Initial scoping experiments included heavier, general-purpose frontier models; 
However, these were excluded from the final macro-evaluation due to prohibitive API costs (scaling to approximately \EUR{5.00} per test subset) and persistent timeout failures during extraction. This instability underscores the fragility of relying on external cloud infrastructure for large-scale archival processing.
Instead, we focus on two alternatives: Mistral-OCR, which provides a highly specialized, geographically sovereign European alternative that processed the dataset for roughly \EUR{2.00}, and Gemini-2.5-Flash, which delivered high-throughput extraction while remaining entirely within the free-tier API limits.

\begin{figure}
    \centering
    \includegraphics[width=.8\textwidth]{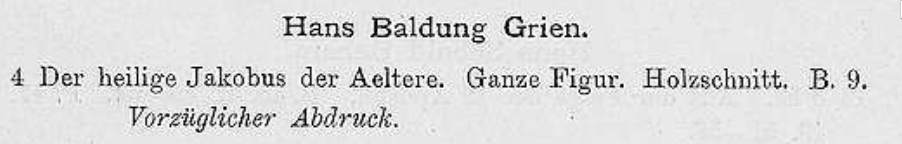}
    \caption{Illustration of an exemplary \ac{ocr} failure from Mistral-OCR. The model falsely included the artist's name in the lot's raw text. Lot from \cite{cat1908}.}
    \label{fig:ocrfail}
\end{figure}

The first finding is that Mistral-OCR has a clear lead in extraction accuracy, achieving an 87\% \ac{anls}* score while still being very affordable. 
% Note that despite our careful revision and manual correction, the results might be biased towards mistral OCRs way of structuring due to our usage of the model for candidate annotations. 
Considering the low 0.03 \ac{cer}, \ac{ocr} failures only partially account for the remaining margin error in \ac{anls}*. 
A failure analysis revealed that these errors boil down to the specific formatting patterns illustrated in \cref{fig:ocrfail}.
Here, the model includes the artist name (Hans Baldung Grien) into the lots' \texttt{raw\_text} field, which leads to a high local \ac{cer} of 0.20.
%Note that despite our careful revision and manual correction of the ground truth, we acknowledge that the comparably high performance of Mistral-OCR might be attributed to the usage of its predictions for the initial candidate annotation.
Note that Mistral's lead should be interpreted in light of the annotation protocol described in \cref{sec:dataset}.
The comparably high performance of Mistral-OCR might be attributed to the usage of its predictions for the initial candidate annotation.

In contrast, Gemini-Flash is considerably weaker (0.75 \ac{anls}*) but faster (19.81 sec/p) and produces negligible costs for datasets of this scale.

For the models provided via AcademicCloud (Mode B), we evaluate the language-focused Gemma4~\cite{team2024gemma} alongside a MoE variant of the vision-focused InternVL3.5~\cite{wang2025internvl3}.
Gemma4-31B achieves higher accuracy than the commercial Gemini-Flash (0.77 \ac{anls}*) but suffers from severe latency (159.34 sec/p), which makes large-scale catalog processing practically infeasible.
Initial attempts to evaluate the latest Qwen3.6~\cite{bai2023qwen} models via AcademicCloud were abandoned entirely due to persistent timeout errors and even slower inference speeds.
Conversely, InternVL3.5~\cite{wang2025internvl3} is the fastest of all evaluated models (17.3 sec/p) but exhibits a slightly lower \ac{anls}* (0.71) and struggles significantly with raw text \ac{ocr}, as evidenced by a high \ac{cer} of 0.21.

When transitioning to locally hosted deployments (Mode C), the 8-bit quantized InternVL3-Q8 variant is considerably weaker (0.61 \ac{anls}*) than the cloud and gateway models, and also markedly slower, at roughly twice the latency of Mistral-OCR (68.30 vs. 30.40 sec/p).
The Qwen3.6-35B-A3B MoE variant performs surprisingly well for a local deployment despite using only 3B active parameters (0.72 \ac{anls}*, 0.16 \ac{cer}), but the slow inference speed (78.96 sec/p) needs to be considered, particularly when large catalog volumes are to be processed. 

Based on these findings, we recommend that institutions with available API budgets and non-sensitive data utilize the highly accurate and cost-effective Mistral-OCR.
However, institutions facing constrained budgets, large data volumes, or strict data privacy regulations can find a viable alternative through institutional gateways hosting mid-sized modern models like Gemma4-31B. 
Even stricter data privacy requirements can be met via the local deployment of quantized models, though this approach strictly requires the use of constrained decoding frameworks. 
Specifically, Mixture of Experts (MoE) variants demonstrate a viable pathway for maintaining adequate performance on restricted hardware. 
Furthermore, our experiments indicate that local deployment without GPU acceleration (or equivalent architectures such as Apple Silicon) is currently not practically viable. 
Finally, regardless of the chosen deployment mode, but particularly when using smaller models, manual verification of the extracted outputs is highly recommended.

\subsection{The Impact of Constrained Decoding}

To isolate the effect of strict schema enforcement on extraction fidelity, we compared the structural accuracy (\ac{anls}*) of models running with constrained decoding against those relying solely on instruction-based prompting (\cref{tab:constrained_decoding}). 

\begin{table}[t]
\centering
\caption{The impact of constrained decoding on structural accuracy (\ac{anls}*).}
\label{tab:constrained_decoding}
\small
\begin{tabular}{lcccc}
\toprule
 & \multicolumn{4}{c}{Evaluated Models (\ac{anls}* $\uparrow$)} \\
\cmidrule(lr){2-5}
Schema Strategy & Gemini-Flash & Mistral-OCR & Gemma4-31B & InternVL3.5 \\
\midrule
Prompt-Based & 0.72 & -- & 0.74 & \textbf{0.71}  \\
Constrained Decoding & \textbf{0.75} & \textbf{0.87} & \textbf{0.77} & 0.69  \\
\bottomrule
\end{tabular}
\end{table}

The experiments show that enforcing schema compliance at the system level leads to a small but consistent improvement in \ac{anls}* for most evaluated architectures. 
A notable exception is the vision-centric InternVL3.5, which exhibited slightly better extraction fidelity (0.71 vs. 0.69) when allowed to generate freely based solely on the system prompt. 
We omit Mistral-OCR from the prompt-based comparison entirely, as its native architectural framework requires structural initialization and does not allow for purely unconstrained generation.

However, while larger models can still function reasonably well using only prompt-based schema adherence, scaling down compute resources alters this dynamic. 
For our locally deployed quantized models (Mode C), purely prompt-based generation entirely failed to produce syntactically valid JSON outputs. 

Overall, these findings suggest that while institutions can successfully rely on prompt-based structuring when utilizing highly capable cloud or institutional gateway models, constrained decoding should be utilized whenever the deployment environment permits it. 
For local edge deployments, it remains an absolute technical requirement.

\subsection{Catalog and Field Analysis}

% First discuss different catalogs:
% aggregate over the three artwork catalogs vs. gemischtwarenladen katalogs 
% add one table similar to table 1 above but reporting only \ac{anls}*(kunst) and \ac{anls}*(misc) 

To understand how catalog composition and field semantics impact extraction fidelity, we disaggregate our evaluation across catalog types and schema keys.

First, we contrast the models' performance on artwork catalogs (1908, 1909, 1931) against the mixed-object catalogs (1932, 1935). 
During dataset curation, we hypothesized that artwork extraction would yield higher accuracy due to its standardized linguistic conventions (e.g., Artist, Title, Medium), compared to the irregular formatting and ambiguous semantic boundaries of mixed household objects.
Surprisingly, the empirical results contradict this assumption.
As shown in \cref{tab:catalog_comparison}, extraction accuracy is systematically lower for art across nearly all evaluated models (with the exception of Gemma4-31B). 
This suggests that the dense, specialized vocabulary and implicit structural dependencies of catalogs offering art pose a greater challenge to \acp{vlm} than the varied but often simpler descriptions of heterogeneous objects.

\begin{table}[t]
\centering
\caption{Catalog-level structural accuracy (\ac{anls}*) comparing artwork catalogs (Art) against mixed-object catalogs (Misc).}
\label{tab:catalog_comparison}
\small
\begin{tabular}{lccccc}
\toprule
Catalog Type & Mistral-OCR & Gemini-Flash & Gemma4-31B & InternVL3.5 & Qwen3.6-35B$^\dagger$ \\
\midrule
\ac{anls}* (Art)  & 0.85      &   0.70     & 0.78     & 0.68  & 0.66 \\
\ac{anls}* (Misc) & 0.91      &   0.83     & 0.75     & 0.79 & 0.79 \\
\bottomrule
\end{tabular}

\vspace{0.5em}
\raggedright
\footnotesize $^\dagger$Mixture of Experts (MoE) variant utilizing 3B active parameters.
\end{table}

% Introduce rouge1 with citation and concise explanation. 
% Add one table with two models (presumably mistral and gemma4 from mode B), reporting rouge-1 and anls* for each field for both.
% Rows are fields.
% Discuss field-based metrics. Which fields work good, which don't? 
% Are there model differences,
While \ac{anls}* provides a rigorous benchmark for exact string and structural compliance, it heavily penalizes minor syntactic deviations via its Levenshtein calculation. 
To evaluate whether lower \ac{anls}* scores indicate true semantic failure or reordering of existing elements, expanding abbreviations etc., 
we introduce \textsc{rouge}-1 \cite{lin2004rouge}. 
By measuring unigram overlap, \textsc{rouge}-1 correctly rewards models that capture the semantic core of an entity, 
even if words are reordered or truncated. 

In \cref{tab:field_analysis}, we compare our highest-performing commercial model (Mistral-OCR) against our most viable institutional alternative (Gemma4-31B) at the field level using both metrics.
Note that the \texttt{lot\_number} field is omitted because our matching strategy based on lot numbers forces a trivial perfect overlap.

\begin{table}[t]
\centering
\caption{Field-level extraction fidelity comparing structural adherence (\ac{anls}*) to semantic recall (\textsc{rouge}-1). The divergence between metrics highlights the subjective nature of free-text fields.}
\label{tab:field_analysis}
\small
\begin{tabular}{lcccc}
\toprule
 & \multicolumn{2}{c}{Mistral-OCR} & \multicolumn{2}{c}{Gemma4-31B} \\
\cmidrule(lr){2-3} \cmidrule(lr){4-5}
Schema Field & \ac{anls}* & \textsc{rouge}-1 & \ac{anls}* & \textsc{rouge}-1 \\
\midrule
\texttt{creator} & 0.81 & 0.94 & 0.76 & 0.88 \\
\texttt{object\_type} & 0.61 & 0.55 & 0.48 & 0.41 \\
\texttt{object\_title} & 0.81 & 0.81 & 0.71 & 0.70  \\
\texttt{place\_of\_creation} & 0.90 & 0.90 & 0.83 & 0.83 \\
\texttt{creation\_time} & 0.94 & 0.94 & 0.88 & 0.89 \\
\texttt{dimensions} & 0.99 & 0.99 & 0.83 & 0.77 \\
\texttt{height} & 0.75  & 0.91 & 0.87 & 0.90 \\
\texttt{width} & 0.75 & 0.91 & 0.89 & 0.90 \\
\texttt{depth} & 0.99 & 0.99 & 0.94 & 0.95 \\
\texttt{weight} & 0.99 & 0.99 & 0.99 & 0.99 \\
\texttt{description} & 0.83 & 0.87 & 0.73 & 0.77 \\
\bottomrule
\end{tabular}
\end{table}

The \texttt{object\_type} field stands out as particularly difficult for both architectures.
This reflects the inherent subjectivity of the category. 
For example, determining whether the ``Barockstuhl'' from \cref{fig:dataset_examples} should be classified broadly as a ``Stuhl'' or retained as the specific composite term frequently leads to misclassifications that penalize both structural and semantic scores.
Conversely, the extraction of spatial measurements (dimensions, depth, weight) generally performs well. 
However, an interesting artifact appears when splitting these measurements into height and width. 
Mistral-OCR exhibits a notable drop in \ac{anls}* (0.75) for these fields while maintaining a high \textsc{rouge}-1 score (0.91). 
A qualitative review of the data reveals that Mistral frequently appends standardized units (e.g., ``cm'') to the raw numerical predictions. 
This stylistic addition triggers a heavy Levenshtein penalty under \ac{anls}*, despite preserving the correct semantic magnitude captured by \textsc{rouge}.

Particularly high discrepancies between the two metrics are also observed in the ``creator'' field (0.81 \ac{anls}* vs. 0.94 \textsc{rouge}-1 for Mistral). 
This is largely explained by mismatches in the ordering of surnames and given names, which is highly inconsistent across historical catalogs and can sometimes even vary within the same document. 
While \acp{vlm} successfully extract the correct name components (rewarded by \textsc{rouge}), they often fail to arrange them in the precise format given by the ground truth. 
This appears as a minor visual error to a human reader. 
But extracting a normalized name representation is critical for subsequent downstream tasks, 
such as disambiguation and connecting to external authority files. 
This highlights that even with the high raw accuracies achieved by state-of-the-art models, a subsequent deterministic post-processing step remains necessary.
Alternatively, improving the prompting strategies, explicitly asking for normalized name representations, or enforcing this via regexp-capable constrained decoding mechanisms might help.

These findings demonstrate that extraction metrics must be contextualized within the definitions of the specific fields and the subjectivity of interpreting historical data.
Establishing a consistent schema with definitions that remain broadly applicable across decades of varying catalog types is difficult. 
Refining these structural guidelines and acknowledging their inherent ambiguities may prove just as impactful for final accuracy metrics as deploying more capable models.

\section{Limitations} 
\label{sec:limitations}

Our model selection was guided by the considerations outlined in \cref{sec:selection}, but it remains non-exhaustive.
The final evaluation was inherently constrained by practical and technical realities, such as persistent timeout errors when accessing the Qwen models via institutional gateways and the prohibitive costs associated with commercial frontier models like Gemini-Pro. 
Given the rapid pace of model development and evolving institutional API provisioning, it is entirely possible that in a few months, the performance measures will be different. 
Nevertheless, we maintain that this evaluation establishes a valuable baseline for the current state of the art in \ac{vlm} access for structured data extraction.
This is particularly relevant for institutions lacking the infrastructure to deploy large models locally, and provides a clear overview of present capabilities.

A second limitation concerns the design of the target schema and the corresponding prompt engineering. 
The semantic mapping of historical auction lots, such as determining the boundaries between physical descriptions, object types, and stylistic epochs, was conducted without directly involving specialized provenance researchers or art historians. 
Consequently, our field definitions and prompt instructions may lack the contextual nuance required by domain experts. 
Also, our current extraction schema cannot model uncertainty and vagueness, which have been argued to be crucial for digital provenance research~\cite{mariani2022introducing}.
Measuring such vagueness could be addressed by annotating the data with multiple annotators and analyzing the inter-annotator agreement.
This would enable to adapt the evaluation scheme to treat ambiguous cases with more flexibility.
Likewise, re-annotating a subset independently of any evaluated system would allow quantifying the bias introduced by our Mistral-OCR-based candidate generation discussed in \cref{sec:dataset}.
Developing evaluation (and potentially also training) strategies that account for these epistemic uncertainties is a promising line of future work. 
To achieve this, we should deepen interdisciplinary collaboration to refine the structural guidelines and ensure that the extracted metadata fully aligns with the epistemic standards of provenance research. 

While beyond the scope of this work, we acknowledge that our evaluation does not include a comparison against traditional, multi-step extraction pipelines. 
Specifically, we did not benchmark the end-to-end \ac{vlm} approach against traditional \ac{ocr} followed by either rule-based regular expression parsing or text-only LLMs. 
Also, state-of-the-art layout understanding models, such as PaddleOCR~\cite{cui2025paddleocr}, were omitted from our comparative baselines.
Furthermore, our current pipeline relies entirely on instruction-based prompting and constrained decoding using the pre-trained weights of the selected Vision-Language Models. 
We anticipate that fine-tuning these models specifically on annotated historical auction records would significantly improve their ability to accurately map complex, domain-specific text to the implicated semantics of our target schema.

We deliberately excluded fine-tuning and complex multi-step pipelines to keep our approach realistic for everyday institutional use. 
Many cultural heritage organizations lack dedicated IT departments or the machine learning expertise required to manage custom training and software orchestration. 
By focusing entirely on out-of-the-box prompting with pre-trained weights, our goal is to provide a pipeline that is immediately actionable and maintainable under current institutional realities.

\section{Conclusion}

In this work, we presented an automated pipeline to extract structured data from historical auction catalogs using \acp{vlm}. 
By turning unstructured page images into searchable data, this approach unlocks new possibilities for large-scale provenance research and art market analysis.

To help cultural heritage institutions navigate real-world technical barriers, we evaluated multiple deployment strategies. 
We found that commercial models like Mistral-OCR offer the highest accuracy and are highly cost-effective. 
However, for institutions with strict privacy needs or limited budgets, locally hosted models or institutional gateways offer a viable alternative, provided they use constrained decoding to strictly enforce the correct output format.

Our field-level analysis also highlighted that historical data is inherently messy and subjective. 
Surprisingly, homogeneous art catalogs that focus on paintings, drawings and prints proved more difficult for the models to process than catalogs offering everyday household items. 
Furthermore, we showed that strict evaluation metrics (like \ac{anls}*) often penalize models for minor formatting differences, even when the models successfully understand the historical meaning of the text.

In future work, we will explore fine-tuning strategies to better capture domain-specific semantics and conduct a deeper comparative analysis to evaluate our pipeline against traditional \ac{ocr} and dedicated layout understanding models. 

Once structured, this data can be integrated into multimodal search engines~\cite{zinnen2026aiding} and linked with other international provenance and art databases.
Ultimately, the successful digitization of historical archives relies not just on advancing computer science, but on close, ongoing collaboration with domain experts to better navigate the complexities of historical records.

Beyond historical auction catalogs, the proposed pipeline and our deployment findings are transferable to similar digitization efforts in museums and collections.
Repurposing this system to convert scans of inventory cards or accession books into structured databases requires only minimal adjustment. 
Updating the target schema and system prompt is expected to be sufficient to adapt to new documentary forms.

\subsubsection{Acknowledgements.}
We thank the Heidelberg University Library for opening up the wealth of historical sales data by providing access to the German Sales catalogs.
Particularly, we would like to thank Maria Effinger and Nicole Sobriel, who were always open to our questions and provided valuable insights.
Furthermore, we would like to extend our thanks to Gabriele Zöllner and the whole SODa team for helpful discussions and feedback.
We also thank AcademicCloud for access to their compute infrastructure, which makes data autonomy and sovereignty accessible to researchers and institutions.
Finally, we thank the anonymous reviewers, whose constructive comments led us to strengthen both the annotation protocol and the evaluation presented here.

This work was partially funded by the German Federal Ministry of Research, Technology and Space (BMFTR) through the SODa project (grant no. 16DKZ2016D).
 
\clearpage
% ============================================================
% Bibliography (ECCV/LNCS: splncs04.bst)
\bibliographystyle{splncs04}
\bibliography{main}

% ============================================================
% BEGIN arXiv-only appendix (content of camera_ready/openreview/supplementary.tex)
% Remove this block to obtain the camera-ready document verbatim.
\clearpage
\appendix
\setcounter{section}{0}
\setcounter{table}{0}
\setcounter{figure}{0}
\renewcommand{\thesection}{S\arabic{section}}
\renewcommand{\thetable}{S\arabic{table}}
\renewcommand{\thefigure}{S\arabic{figure}}
\renewcommand{\theequation}{S\arabic{equation}}

\section*{Supplementary Material}

\section{Evaluation Including Unmatched Lots}
\label{sec:unmatched_lots}

As noted in the main manuscript, historical auction catalogs frequently contain implicit hierarchical structures that \acp{vlm} may parse as distinct physical entities. While these structural over-generations (hallucinated lots) are reliably filtered out via deterministic post-processing in our primary evaluation, this section presents the raw, unfiltered performance metrics. \cref{tab:supp_unmatched} reports the pipeline's performance when unmatched lots are retained, meaning both \ac{anls}* and \ac{cer} strictly penalize the generation of hallucinated or unassigned lot data prior to any heuristic intervention.

\begin{table}[h]
\centering
\caption{Macro-level comparison of model configurations evaluated, including unmatched (hallucinated) lots. Unlike the primary evaluation, the structural accuracy (ANLS*) and transcription fidelity (CER) metrics reported here penalize structural over-generation. Modes represent commercial cloud APIs (A), institutional gateways (B), and local edge deployments (C).}
\label{tab:supp_unmatched}
\small
\begin{tabular}{clccc}
\toprule
Mode & Model &   ANLS* ($\uparrow$) & CER ($\downarrow$) & sec/p ($\downarrow$) \\
\midrule
\multirow{2}{*}{A} & Gemini-Flash  & 0.59 & 0.25 & 19.81 \\
& Mistral-OCR  & \textbf{0.69}  & \textbf{0.03} & 30.4 \\
\midrule
\multirow{2}{*}{B} & Gemma4-31B & 0.69 & 0.21 & 159.34 \\
& InternVL3.5-30B-A3B & 0.64 & 0.29 & \textbf{17.3} \\
\midrule
\multirow{2}{*}{C} & InternVL3-Q8 & 0.51 & 0.36 & 68.3 \\
& Qwen3.6-35B-A3B & 0.50 & 0.38 & 78.96 \\
\bottomrule
\end{tabular}
\end{table}
% END arXiv-only appendix
% ============================================================

\end{document}

%% file: figs/overview3modes.tex
\begin{tikzpicture}[
        % Define global styles for consistent aesthetics
        base/.style={align=center, minimum height=1.2cm, draw=black!70, thick, rounded corners=3pt, font=\sffamily\small},
        document/.style={base, text width=2.2cm, fill=gray!10, minimum height=3cm},
        mode/.style={base, text width=3cm, fill=blue!5, minimum height=0.8cm},
        constraint/.style={base, text width=2.2cm, fill=orange!10, minimum height=2cm, dashed},
        output/.style={base, text width=2.2cm, fill=green!10, minimum height=2cm, font=\ttfamily\scriptsize},
        downstream/.style={base, text width=3cm, fill=purple!10},
        arrow/.style={thick, color=black!70, -{Stealth[scale=1.2]}},
        groupbox/.style={draw=black!30, dashed, inner sep=10pt, rounded corners}
    ]

    % --- 1. THE INPUT ---
    \node[document] (input) {Historical\\Catalog Page\\[1em] \textit{(Unstructured layout)}};

    % --- 2. THE DEPLOYMENT MODES (Engine) ---
    % Positioned relative to the input
    \node[mode, right=1.5cm of input, yshift=1.2cm] (modeA) {\textbf{Mode A}\\Commercial APIs};
    \node[mode, below=0.3cm of modeA] (modeB) {\textbf{Mode B}\\Institutional APIs};
    \node[mode, below=0.3cm of modeB] (modeC) {\textbf{Mode C}\\Local Deployment};
    % \node[mode, below=0.3cm of modeC] (modeD) {\textbf{Mode D}\\Consumer Local};

    % Draw a background box around the modes to group them
    \begin{scope}[on background layer]
        \node[groupbox, fit=(modeA)(modeB)(modeC), label={[font=\sffamily\bfseries\small, text=black!70]below:Deployment Modes}] (modes_box) {};
    \end{scope}

    % --- Establish the vertical center point ---
    % This creates an invisible coordinate 2cm to the right of the exact center of modes_box
    \coordinate (right_center) at ($(modes_box.east) + (2.5cm, 0)$);

    % --- 3. THE FILTER (Constraints) ---
    % Placed just above the center line
    \node[constraint, above=0.25cm of right_center] (filter) {\textbf{Constrained\\Decoding}\\[0.5em] };

    % --- 4. THE OUTPUT ---
    % Placed just below the center line (creating a perfect 0.5cm gap between them)
    \node[output, below=0.25cm of right_center] (json) {{\{}\\ "lot": "12",\\ "artist": "...",\\ "price": "..."\\ {\}}};

    % --- 5. DOWNSTREAM APPLICATION --- CAN be removed if we need space
    \node[downstream, below=0.7cm of input] (cbir) {\textbf{Multimodal\\Search \& CBIR}};

    % --- Establish the vertical center point ---
    % This creates an invisible coordinate 2cm to the right of the exact center of modes_box
    \coordinate (right_center) at ($(modes_box.east) + (2cm, 0)$);

    %% ARROWS

    % Connect Input to Modes Box
    \draw[arrow] (input.east) -- (input.east -| modes_box.west) node[align=center,midway, below, font=\sffamily\scriptsize] {VLM \\ Inference};
    % Place "Enforced" on the left of the line, and "Schema" on the right

    % Connect Modes Box to Filter
    % We route from the center of the modes box to the left side of the filter
    \draw[arrow] (modes_box) -- (filter.west);

    \draw[arrow] (modes_box) -- (json.west) node[align=center,midway,below=0.2cm,font=\sffamily\scriptsize] {Prompted \\ Schema};

    % Connect Filter to JSON
    % \draw[arrow] (filter.south) -- (json.north) node[midway, right, font=\sffamily\scriptsize] {Enforced Schema};
    \draw[arrow] (filter.south) -- (json.north) 
        node[midway, left, font=\sffamily\scriptsize] {Enforced}
        node[midway, right, font=\sffamily\scriptsize] {Schema};

    % Connect JSON to Downstream
    % THis can be removed if we need space
    \draw[arrow] (json.south) |- (cbir.east) node[align=center,pos=0.60,below,font=\sffamily\scriptsize] {Future Applications};

    \end{tikzpicture}